\documentclass[manuscript,screen,nonacm]{acmart}

\renewcommand\footnotetextcopyrightpermission[1]{} 
\usepackage{fancyhdr}
\fancypagestyle{plain}{
  \fancyhf{} 
  \fancyfoot[C]{\textit{Manuscript – Preprint Version}} 
}
\setcopyright{none}

\usepackage{xcolor}
\usepackage{enumitem}   
\usepackage{multirow}
\usepackage{array}

\newcommand\Vcline[1]{%
  \noalign{\vskip\arrayrulewidth\global\let\CT@do@color\relax}%
  \cline{#1}%
  \noalign{\vskip\arrayrulewidth}
}
\makeatother
\usepackage{float}
\usepackage{adjustbox}
\usepackage{booktabs}
\usepackage{lipsum}         
\usepackage{booktabs}       
\usepackage{adjustbox}      
\usepackage{changepage}     
\usepackage{graphicx} %

\usepackage{adjustbox}
\usepackage{arydshln}

\usepackage{amssymb}

\usepackage{amsmath} 
\usepackage{bm} 
\usepackage{algorithm}
\usepackage{algpseudocode}
\algnewcommand{\LeftComment}[1]{\Statex \(\triangleright\) #1}

\newcommand{\blue}[1]{\textit{\color{blue} #1}}
\newcommand{\green}[1]{\textbf{\color{green} #1}}

\newcommand{\red}[1]{\textbf{\color{red} #1}}
\newcommand{\olive}[1]{\textbf{\color{olive} #1}}

\begin{document}

\title{Low-Resource Neural Machine Translation Using Recurrent Neural Networks and Transfer Learning: A Case Study on English-to-Igbo}

\author{Ocheme Anthony Ekle}
\authornote{Both authors contributed equally to this research.}
\authornote{This research was conducted while the author was at \textbf{Moscow Institute of Physics and Technology (MIPT)}, Russia.}
\authornote{Author is currently affiliated with \textbf{Tennessee Technological University}, USA.}
\email{oaekle42@tntech.edu} 
\orcid{https://orcid.org/0009-0003-6204-0657}
\email{ekleanthony@phystech.edu}
\affiliation{%
  \institution{Tennessee Technological University}
  \streetaddress{1020 Stadium Drive, 406.}
  \city{Cookeville}
  \state{TN}
  \country{USA}
  \postcode{38505}
}

\author{Biswarup Das}
\orcid{https://orcid.org/0000-0002-6133-9820}
\authornotemark[1]
\authornotemark[2]
\affiliation{%
  \institution{Moscow Institute of Physics and Technology }
  \authornotemark[1]
  \streetaddress{ 9 Institutskiy pereulok,}
  \city{Dolgoprudny, Moscow region}
  \country{Russia}}
\email{biswarup.das1@huawei.com}


\renewcommand{\shortauthors}{Ekle O.A, et al.}

\begin{abstract}
  In this study, we develop Neural Machine Translation (NMT) and Transformer-based transfer learning models for English-to-Igbo translation—a low-resource African language spoken by over 40 million people across Nigeria and West Africa. Our models are trained on a curated and benchmarked dataset compiled from Bible corpora, local news, Wikipedia articles, and Common Crawl, all verified by native language experts. We leverage Recurrent Neural Network (RNN) architectures, including Long Short-Term Memory (LSTM) and Gated Recurrent Units (GRU), enhanced with attention mechanisms to improve translation accuracy. To further enhance performance, we apply transfer learning using MarianNMT pre-trained models within the SimpleTransformers framework. Our RNN-based system achieves competitive results, closely matching existing English-Igbo benchmarks. With transfer learning, we observe a performance gain of +4.83 BLEU points, reaching an estimated translation accuracy of 70\%. These findings highlight the effectiveness of combining RNNs with transfer learning to address the performance gap in low-resource language translation tasks.
\end{abstract}




\maketitle

\section{Introduction}
\label{sec:introduction}

The growing need for advanced translation systems and the increasing linguistic diversity across the globe have posed significant challenges in translating text from one language to another. This demand has led to significant research in the field of Natural Language Processing (NLP), particularly in Machine Translation (MT)~\cite{01_Priya2021_analysis_nlp}. Machine translation is a key application of NLP, involving automated systems that translate text between languages. The primary goal of machine translation is to ensure that the translated sentence in the target language conveys the same meaning as the original sentence in the source language~\cite{02_sutskever2014}.

Past researchers have adopted a variety of approaches for building machine translation systems, including Rule-Based Machine Translation (RBMT), Corpus-Based or Statistical Machine Translation (SMT), and Hybrid Machine Translation (HMT), which combines elements of both RBMT and SMT. These paradigms encompass word-based, phrase-based, forest-based, and more recently, Neural Machine Translation (NMT) methods~\cite{04_okpor2014MT}.

Neural Machine Translation has emerged as a significant breakthrough in the MT landscape, leveraging deep neural networks and large language models \cite{qin2024large_language_model_survey} to model translation tasks more effectively~\cite{02_sutskever2014}. Typically, NMT systems adopt an encoder-decoder architecture \cite{02_sutskever2014}, where the encoder processes the input sentence and the decoder generates the translated output~\cite{05_Costa2015_hybrid_mt}. These architectures, especially when enhanced with attention mechanisms, have demonstrated state-of-the-art performance and significantly outperformed earlier translation techniques~\cite{06_bahdanau2014}.

Studies have shown that NMT models not only learn faster but also produce more fluent and accurate translations compared to traditional methods. Recurrent Neural Network (RNN)-based models, particularly those using Long Short-Term Memory (LSTM) and Gated Recurrent Unit (GRU) architectures, are widely used due to their ability to capture sequential dependencies in text. However, despite the progress in NMT, limited research has been directed towards developing systems for low-resource and endangered languages—especially African languages~\cite{03_ezeani2020igbo}. To address this gap, our research focuses on building an NMT model for translating English into Igbo, a major African language spoken by over 40 million people across Nigeria and West Africa.

Our key contributions are summarized as follows:
\begin{itemize}
    \item We present the first high-performing translation system for the low-resource English–Igbo language pair, incorporating attention mechanisms~\cite{06_bahdanau2014} and global attention~\cite{10_luong2015_attention_paper}.
    
    \item We integrate the teacher forcing algorithm to provide the decoder with accurate reference outputs during training, thereby improving translation quality.
    
    \item We implement and compare both Greedy Decoding and Beam Search strategies to enhance translation fluency and accuracy.
    
    \item We conduct extensive experiments using approximately 12,000 parallel sentence pairs. Our transformer-based transfer learning model~\cite{19_zoph2016transfer_learning_lowRes, 20_nguyen2017transfer, 21_nair2022transfer} achieves a peak translation accuracy of \textbf{70\%} and demonstrates a notable improvement of \textbf{+4.83 BLEU} points over existing benchmarks for English–Igbo translation.
\end{itemize}

The remainder of this paper is organized as follows: Section~\ref{sec:related_works} reviews related literature in NLP and NMT. Section~\ref{sec:backgroud_and_rnn_review} outlines the foundational concepts of RNNs and discusses their limitations. Section~\ref{sec:proposed_method} presents the architecture of our proposed translation model. Section~\ref{sec:experiments_results} details the experimental setup, including comparisons of RNN variants, attention mechanisms, decoding strategies, datasets, and the application of transfer learning. Section~\ref{sec:discussion} presents quantitative and qualitative results with performance evaluations. Finally, Sections~\ref{sec:discussion} and~\ref{sec:conclusion} offer a comprehensive discussion, conclusions, identified limitations, and future research directions.

\section{Related Work}
\label{sec:related_works}
This section provides a review of the literature on Natural Language Processing (NLP), levels of language understanding, word representation, and machine translation. It also highlights recent developments in graph-based learning and generative models, showing how they relate to low-resource language translation.

\subsection{Natural Language Processing and Language Levels}
\label{sec:nlp_language_level}

Research in NLP began in the late 1940s, combining concepts from artificial intelligence (AI) and linguistics, with machine translation (MT) as one of its earliest applications \cite{nadkarni2011nlp, liddy2001nlp}. In 1947, Warren Weaver proposed using cryptography and information theory for language translation \cite{schwartz2018mt_history}; however, his approach struggled with lexical ambiguity. This challenge inspired deeper exploration in the field. Noam Chomsky’s landmark 1957 work, \textit{Syntactic Structures}, introduced generative grammar, significantly shaping NLP and MT \cite{lees1957syntactic}. This era also saw early efforts in speech recognition, heuristic reasoning, and question answering. Between the 1970s and 1990s, NLP advanced with innovations like semantic parsing and discourse modeling \cite{kumar2011nlp}. Schank and Tesler developed language understanding programs focused on human conceptual knowledge, such as memory and goals \cite{roger_schank1969}.

Since the 1990s, probabilistic and data-driven methods have dominated NLP, with models like probabilistic parsing and word embeddings. Advances in computing hardware have further enhanced speech and language processing capabilities \cite{liddy2001nlp}. Today, researchers focus on next-generation applications that better handle linguistic variability and ambiguity. A core objective is to design computational techniques that learn and represent text data in a human-like manner. Converting symbolic text into numerical form remains a foundational step in NLP.

\textbf{Language Levels:} NLP operates across multiple linguistic levels: phonology, morphology, lexical, syntactic, semantic, pragmatic, and discourse \cite{kumar2011nlp}. These layers are interdependent and essential for understanding and generating language. Ambiguity remains a major challenge, addressed using knowledge-based methods (e.g., Mahesh and Nirenburg, 1996) and modern techniques like FastText \cite{uslu2018}. \textbf{Phonology} deals with sound patterns; \textbf{morphology} analyzes word structure; \textbf{syntax} examines sentence formation; \textbf{lexicon} interprets word meanings; \textbf{semantics} studies meaning within phrases; \textbf{discourse} handles coherence across text units; and \textbf{pragmatics} evaluates intent and contextual meaning \cite{liddy2001nlp}.

\subsection{Word Vector Representation}
\label{sec:word_vec_representation}

Word vector representations are crucial in NLP for capturing the semantic relationships between words \cite{brownleeOneHot}. Early methods used “one-hot” encodings, which lacked the ability to capture similarity between words. Firth’s idea of representing words by their context laid the foundation for embeddings. Bengio et al. \cite{bengio2003} introduced neural word embeddings, followed by Collobert and Weston \cite{collobert2008unified}, who applied multitask and semi-supervised learning.

Mikolov et al. \cite{mikolov2013linguistic} introduced Word2Vec, featuring the continuous bag-of-words (CBOW) and skip-gram architectures. CBOW predicts a word given its context, while skip-gram predicts context from a given word. GloVe, proposed by Pennington et al. \cite{pennington2014glove}, is another widely used pretrained embedding. Facebook’s FastText \cite{bojanowski2017enriching_fasttext} captures subword information and polysemy, and later work incorporated uncertainty modeling \cite{athiwaratkun2018probabilistic}.

More recently, contextualized word embeddings such as BERT \cite{devlin2018bert}, ELMo, and GPT-2 \cite{ethayarajh2019contextual} have surpassed older models, outperforming both one-hot and bag-of-words methods \cite{palash_goyal2018}. These embeddings are now central to both supervised and unsupervised tasks, powering neural machine translation, question answering, and information retrieval.

In this study, we employ the Keras tokenizer to convert text into numerical sequences using morphology-based tokenization. We also use pretrained word embeddings via Keras’ embedding layer to generate dense vector representations.

\subsection{Neural Machine Translation}
\label{sec:nmt_review}

The origin of MT dates back to Warren Weaver’s 1946 proposal of using cryptographic techniques for language translation \cite{07_hutchins1997}. In later years, researchers began combining rule-based and statistical methods, giving rise to hybrid models for neural machine translation (NMT) \cite{08_yamada2001_syntax_based_stat_mt}.

Modern NMT systems typically follow the sequence-to-sequence (Seq2Seq) paradigm, composed of an encoder and decoder \cite{02_sutskever2014}. The encoder transforms the input sentence into a fixed-length vector (hidden state), while the decoder generates the output sequence based on this vector. However, traditional encoder-decoder models face two primary challenges:

\textbf{(1)} Long-distance dependencies between words can degrade translation accuracy. Gusev and Oboturov addressed this by reversing the encoder direction to assist training.  
\textbf{(2)} Compressing the entire input sequence into a single vector introduces an \textbf{information bottleneck}, especially for long inputs.

Bahdanau et al. \cite{06_bahdanau2014} addressed these challenges using an attention mechanism. Rather than relying on a single fixed vector, attention allows the decoder to reference all encoder states dynamically at each step, focusing on relevant input tokens during decoding. This greatly improved translation quality, but the method increased computational complexity and still struggled with low-resource data and long-term dependencies.

Luong et al. \cite{10_luong2015_attention_paper} later introduced local attention, improving alignment between source and target tokens. Self-attention networks, such as those developed by Yang et al. \cite{11_yang2019cnn_selfattention} and Shaw et al. \cite{22_shaw2018self_atten_relative_postion}, further advanced sequence modeling by enabling tokens to attend to different positions within a sentence.

Besides RNN-based models like LSTM and GRU, CNNs have also been explored for NMT. Kalchbrenner et al. \cite{13_kalchbrenner2016nmt_lineartime} and Bradbury et al. \cite{12_bradbury2016quasi_rnn} applied CNNs for efficient sequence modeling, while Gehring et al. \cite{14_gehring2017cnn_seq} used CNNs with attention to model long dependencies. Other innovations include Connectionist Temporal Classification (CTC) by Graves et al. \cite{15_graves2006connectionist} and model ensembling for robust predictions \cite{16_neubig2017nmt_seq}.

A breakthrough in NMT came with the Transformer architecture proposed by Vaswani et al. \cite{17_vaswani2017attention_is_all_you_need}. This model replaced recurrence with self-attention, enabling parallelization and improved performance. Key components include multi-head attention and position-wise feed-forward networks. Transformers laid the foundation for models like GPT-2 \cite{ethayarajh2019contextual}, BERT \cite{gregory2008mir_BERT}, and T5 \cite{raffel2020exploring_T5_text}.

Despite their success, Transformer-based models often underperform on low-resource languages \cite{19_zoph2016transfer_learning_lowRes}. To address this, transfer learning has been proposed: a high-resource (parent) model is trained first, and its knowledge is transferred to a low-resource (child) model \cite{19_zoph2016transfer_learning_lowRes, 20_nguyen2017transfer, 21_nair2022transfer}. This approach improves performance by leveraging shared representations.

\subsection{Graph Neural Networks and Emerging Trends in NLP}

While traditional NLP models rely on sequential or convolutional architectures, recent research has shown that Graph Neural Networks (GNNs) \cite{bastings2017GNN} can effectively model structured relationships in linguistic data, such as syntactic and semantic dependencies \cite{04_sp_waikhom2021GNN_Methos_shallow_deep_embedding}. GNNs have been applied in tasks like word embedding \cite{ekle2024_dynamicGraph2024}, knowledge graph completion, and low-resource machine translation by capturing long-range interactions across words and phrases in a graph structure \cite{msc_1_liu2022resource_GNN, ekle2024_dynamicGraph2024}. For instance, in translation tasks, representing words as nodes and their co-occurrence or syntactic relations as edges enables the model to preserve richer context.

In parallel research, GNN-based techniques have also proven effective for anomaly detection in dynamic graph streams \cite{043_F-FADE_chang2021f, 070_midas_bhatia2020midas, ekle2024_DecayRAnk, AdaptiveDecayRank_ekle2025adaptive}. These contributions highlight the growing role of GNNs in both structured NLP and dynamic graph-based learning.

Furthermore, the emergence of Large Language Models (LLMs), such as GPT, multilingual BERT variants, and recent generative embedding techniques, has pushed the boundaries of translation and multilingual understanding. These models leverage large-scale pretraining to capture context-rich representations and enable robust performance in low-resource settings \cite{brown2020gpt3, xue2021mt5}. Generative embeddings are increasingly being used in cross-lingual retrieval, summarization, and neural translation \cite{reimers2022sbert, wang2023genembedding}, offering new opportunities for integrating contextual, graph-based, and generative learning.

In this research, we focus on English-to-Igbo translation—a low-resource language pair. We adopt the RNN-based encoder-decoder framework and enhance it with attention and global attention mechanisms \cite{06_bahdanau2014, 10_luong2015_attention_paper}. We also apply the teacher forcing algorithm \cite{lamb2016professor_teacherforcing} to guide decoding during training and incorporate transfer learning to optimize performance.

\textbf{Goal:} By leveraging these techniques, we aim to build an effective neural machine translation model that bridges the language gap between English and Igbo, contributing to the development of low-resource NLP applications.
\section{Background and RNN Architectures}
\label{sec:backgroud_and_rnn_review}
This section provides the foundational background for Recurrent Neural Networks (RNNs), including the classical RNN architecture and its challenges. We also explore advanced variants such as Long Short-Term Memory (LSTM) and Gated Recurrent Units (GRU). Furthermore, we introduce the sequence-to-sequence (Seq2Seq) encoder-decoder model, highlight its limitations, and describe how the attention mechanism addresses those challenges.

\subsection{Traditional RNNs and the Vanishing Gradient Problem}
\label{sec:traditional_RNN}

Recurrent Neural Networks (RNNs) are designed to model temporal or sequential data by maintaining hidden states that capture information from previous time steps. This ability makes them suitable for a variety of time-dependent tasks, including financial forecasting \cite{giles1997}, electric load prediction \cite{costa1999LSTM}, and water quality monitoring \cite{jain_medsker2000rnn}.

Early RNN architectures, such as the Vanilla RNN, Elman Network, and Jordan Network \cite{Elman_RNN_pham1999training}, encountered limitations due to the vanishing and exploding gradient problems. These issues arise during backpropagation through time (BPTT), making it difficult for the network to learn long-term dependencies \cite{debabala2019, palash_goyal2018}. As a result, traditional RNNs struggle when the number of time steps increases significantly.

\begin{figure}[h]
  \centering
  \includegraphics[width=0.65\linewidth]{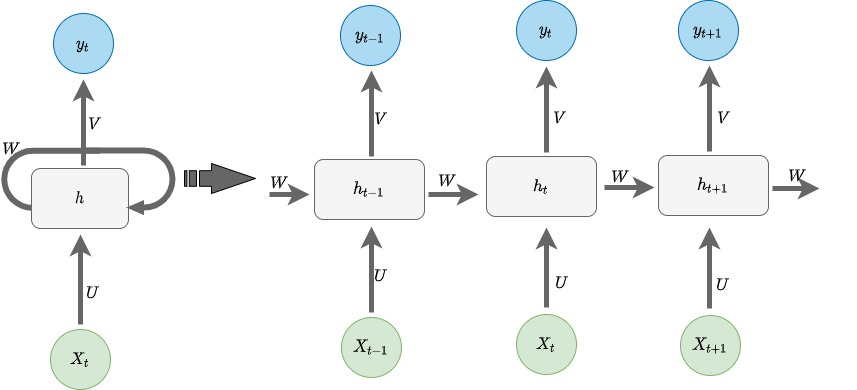}
  \caption{\textbf{Vanilla RNN architecture.} The weight matrices $U$, $V$, and $W$ are shared across time steps to process temporal sequences.}
  \label{fig:vanilla_rnn}
  \Description{The Vanilla RNN Architecture showing input, hidden, and output connections over time.}
\end{figure}

Figure~\ref{fig:vanilla_rnn} illustrates the architecture of a Vanilla RNN, where the same set of weights ($U$, $V$, and $W$) are used across time steps to process sequential data.

To overcome these challenges, more advanced architectures such as Long Short-Term Memory (LSTM) and Gated Recurrent Units (GRU) were introduced \cite{heck2017simplified}. Another notable variant is the Continuous-Time RNN (CTRNN), used in adaptive behavior modeling and robotics \cite{beer1997}. These improvements allow RNN-based models to learn temporal dependencies over longer sequences more effectively.

\subsection{Long Short-Term Memory (LSTM)}
\label{sec:lstm}

LSTM networks, proposed by Hochreiter and Schmidhuber \cite{hochreiter1997lstm}, are designed to address the vanishing gradient problem inherent in traditional RNNs. LSTM introduces gated mechanisms—namely, the \textit{input gate}, \textit{forget gate}, and \textit{output gate}—to regulate the flow of information through a memory cell \cite{felix_gers2002lstm}. This enables the network to retain or discard information over extended time intervals, making it well-suited for modeling long-term dependencies.

LSTMs have achieved state-of-the-art results in several domains, including handwriting recognition and speech processing \cite{schmidhuber2015}. They have also been successfully trained using the Connectionist Temporal Classification (CTC) loss function for tasks requiring flexible sequence alignment \cite{15_graves2006connectionist}. According to Xuan Le \cite{xuan_le2019application}, LSTM-based models have set benchmarks in international pattern recognition competitions.

\subsection{Gated Recurrent Unit (GRU)}
\label{sec:gru}

The Gated Recurrent Unit (GRU), introduced by Cho et al. in 2014 \cite{cho2014rnn_encdec}, is a streamlined alternative to LSTM. It combines the functionalities of the input and forget gates into a single \textit{update gate}, and uses a \textit{reset gate} to control the contribution of previous hidden states. With fewer parameters, GRUs offer computational efficiency while maintaining comparable performance to LSTMs in many NLP applications \cite{heck2017simplified, zhao2019evaluation, zhou2016minimal_gru}.

\subsection{Summary and Implementation}
\label{sec:rnn_summary}

In this study, we implemented both LSTM and GRU architectures within an encoder-decoder sequence-to-sequence framework. To enhance translation accuracy, we integrated the attention mechanism, which helps the model focus on relevant parts of the input during decoding. Additionally, we employed the teacher-forcing algorithm to guide the decoder using the ground truth output during training. These techniques collectively improve the model’s ability to handle long-range dependencies and produce higher-quality translations for low-resource language pairs.

\section{PROPOSED METHOD }
\label{sec:proposed_method}

This section outlines the complete pipeline and methodology adopted to build and evaluate our low-resource English-to-Igbo neural machine translation (NMT) system. We provide comprehensive explanations of the data acquisition, preprocessing, model architecture, training and inference mechanisms, transfer learning integration, and evaluation strategies. We leverage Recurrent Neural Networks (RNNs) with attention and explore how Transformer-based pretrained models can further boost translation performance.

\subsection{Data Collection}
Approximately 12,000 English-Igbo parallel corpus sentences were extracted from the \href{https://github.com/IgnatiusEzeani/IGBONLP/tree/master/ig_en_mt/benchmark_dataset}{GitHub repository} curated by \cite{03_ezeani2020igbo}. This benchmark dataset comprises verified translations from Bible corpora, local news sources, Wikipedia articles, Common Crawl data, and various indigenous materials. Linguistic experts reviewed and cleaned the data to ensure alignment consistency and language accuracy. For comparative benchmarking and transfer learning exploration, we also evaluated our model using the English-French (Eng-Fra) dataset from the Tatoeba corpus.

\begin{figure}[hbt!]
\centering
\includegraphics[scale=0.3]{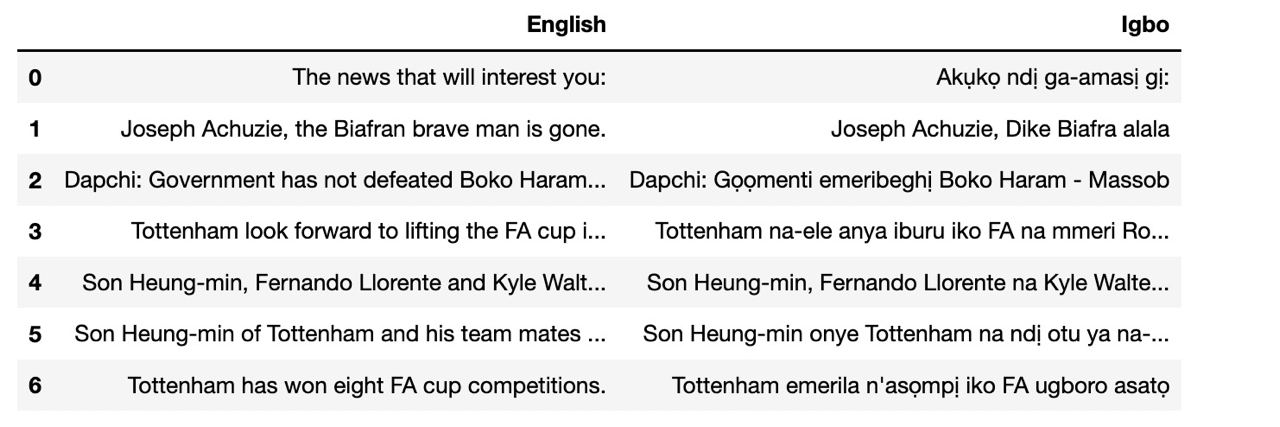}
\caption{Sample English-Igbo Dataset Visualization}
\label{fig:sample_data}
\end{figure}

\begin{figure}[hbt!]
\centering
\includegraphics[scale=0.5]{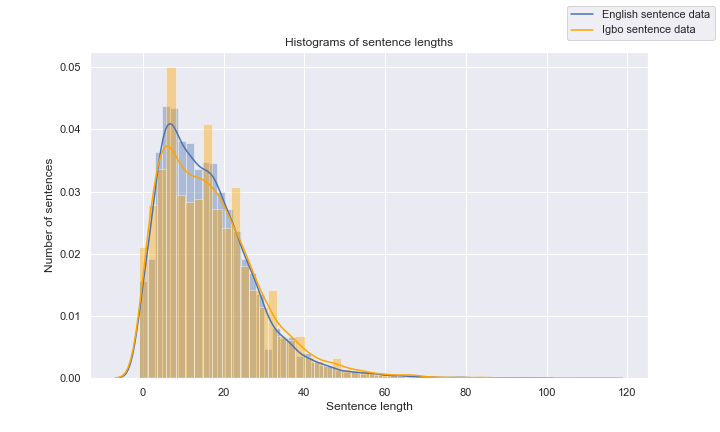}
\caption{Histogram of Sentence Lengths}
\label{fig:histogram}
\end{figure}

\subsection{Text Preprocessing and Analysis}
To ensure the raw parallel corpus was ready for training, we implemented a three-stage preprocessing pipeline: data loading and visualization, tokenization, and vocabulary construction.

\subsubsection{Data Loading and Sampling}
The parallel corpus was loaded into memory, where each English-Igbo sentence pair was separated by a tab. These sentence pairs were visualized using pandas DataFrames to verify the consistency and format, as shown in Figure~\ref{fig:sample_data}. This initial step helped ensure that sentence alignments were correct and suitable for downstream tokenization and model input preparation.

\subsubsection{Sentence Length Analysis}
\begin{figure}[hbt!]
\centering
\includegraphics[scale=0.3]{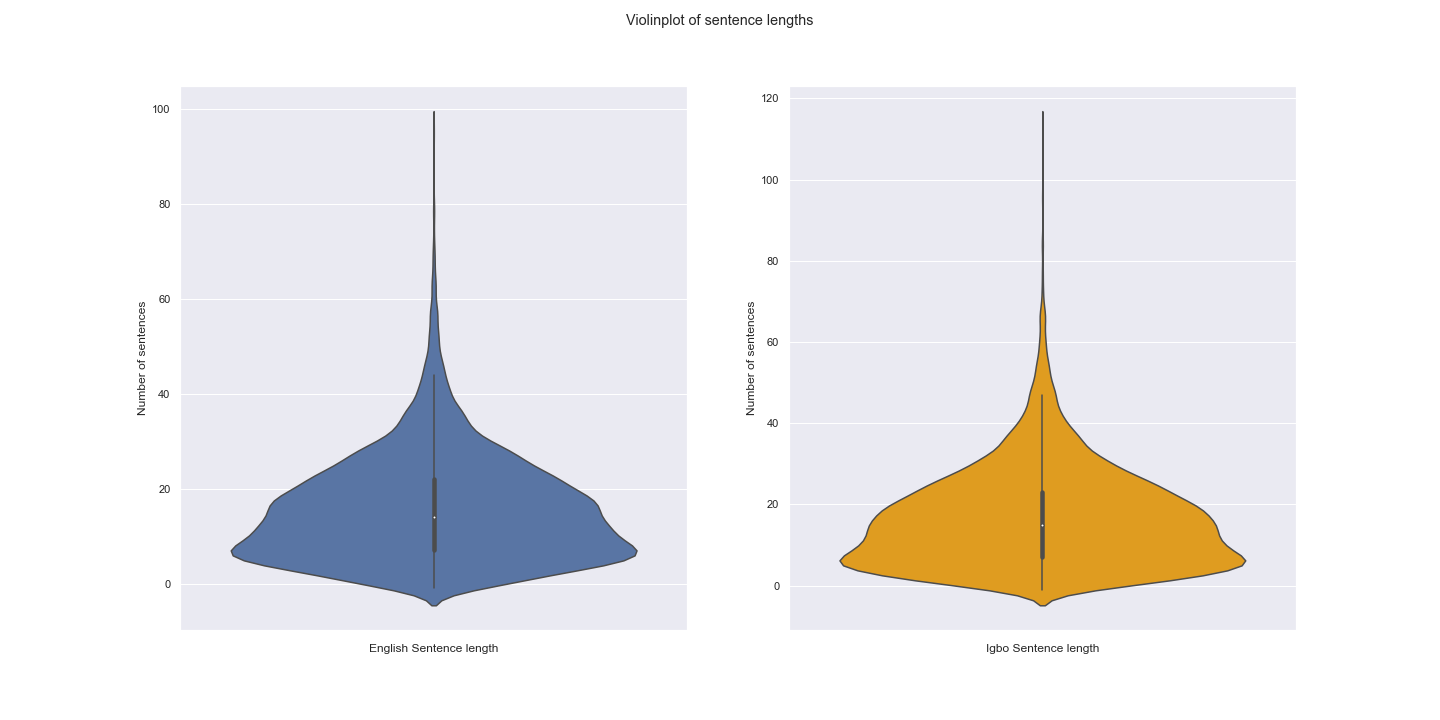}
\caption{Violin Plot of Sentence Lengths}
\label{fig:violin_plot}
\end{figure}

We conducted Exploratory Data Analysis (EDA) to assess the distribution of sentence lengths. Understanding sentence lengths is crucial for configuring the input sequence padding and optimizing memory usage. As illustrated in Figures~\ref{fig:histogram} and \ref{fig:violin_plot}, most sentence lengths range between 10 and 40 tokens. These insights informed the padding strategy and helped us avoid excessive memory consumption or model truncation.

\subsubsection{Tokenization and Vocabulary Building}
We used the Keras Tokenizer to perform morphology-aware tokenization. Punctuation and digits were filtered out, and all text was lowercased. Tokens were mapped to unique indices to build separate vocabularies for English and Igbo. Special tokens \texttt{<pad>}, \texttt{<sos>}, and \texttt{<eos>} were added to denote padding, sentence start, and sentence end, respectively. Table~\ref{tab:vocabulary} presents the vocabulary size and word count.

\begin{table}[hbt!]
\centering
\begin{tabular}{lcc}
\toprule
 & English & Igbo \\
\midrule
Total Words & 176,375 & 184,538 \\
Vocabulary Size & 16,224 & 14,789 \\
\bottomrule
\end{tabular}
\caption{Vocabulary Statistics}
\label{tab:vocabulary}
\end{table}

\subsection{Model Design: RNN with Attention}
\begin{figure}[hbt!]
\centering
\includegraphics[scale=0.4]{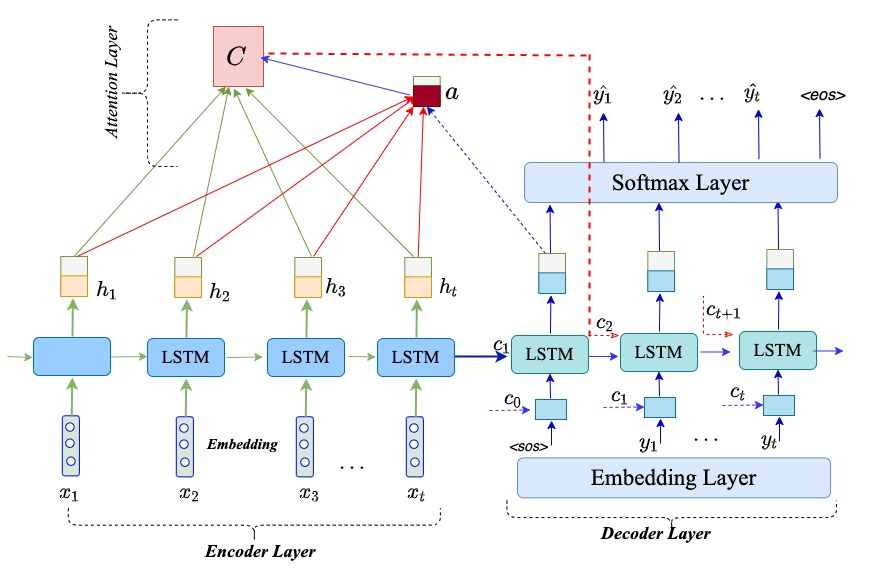}
\caption{RNN Encoder-Decoder Model with Attention}
\label{fig:seq2seq_model}
\end{figure}

The proposed model follows the classical sequence-to-sequence (Seq2Seq) framework enhanced with attention. The architecture comprises three key components: an encoder, a decoder, and an attention layer in between. The overall architecture is shown in Figure~\ref{fig:seq2seq_model}.

This modular structure enables the encoder to process and compress input sequences into hidden representations, which the attention-enhanced decoder uses to generate target translations.

\subsubsection{Encoder}
The encoder uses an embedding layer followed by Long Short-Term Memory (LSTM) units to process input sequences. At each time step $t$, the encoder computes the hidden state $h_t$ and updates the context vector $C$:
\begin{equation}
  h_t = \sigma_1(x_t, h_{t-1}), \quad C = \sigma_2(h_1, \dots, h_t)
\end{equation}
where $\sigma_1$ and $\sigma_2$ are non-linear activation functions. We configured the encoder with a batch size of 128, embedding dimension of 256, and 1024 LSTM units. These hyperparameters were selected based on validation performance and training stability.

\subsubsection{Decoder with Attention}
The decoder mirrors the encoder’s structure and integrates Bahdanau-style attention \cite{06_bahdanau2014}. At each decoding step, the decoder uses the previous output and attention to generate the next word. The attention mechanism computes a context vector $C_t$ as:
\begin{equation}
  C_t = \sum_j a_{tj} h_j
\end{equation}
where
\begin{equation}
  a_{tj} = \frac{\exp(e_{tj})}{\sum_k \exp(e_{tk})}, \quad e_{tj} = \text{score}(h_{t-1}^{\text{dec}}, h_j^{\text{enc}})
\end{equation}
We experimented with dot, general, concat, and scaled dot-product attention. We adopted global attention \citep{10_luong2015_attention_paper} for its balance between interpretability and performance.

\subsubsection{Activation Functions and Loss}
The encoder and decoder utilized Sigmoid, Tanh, and Softmax activation functions. The final output layer applied Softmax to produce a probability distribution. We used sparse Softmax cross-entropy loss:
\begin{equation}
  \mathcal{L}(y, f) = -\sum_{i=1}^C y_i \log\left(\frac{e^{f_y}}{\sum_{j=1}^C e^{f_j}}\right)
\end{equation}
This loss function is suitable for multiclass classification and provides a stable gradient during training.

\subsubsection{Optimization}
The Adam optimizer \citep{kingma2014adam} was employed with a learning rate of 0.001. Adam dynamically adjusts learning rates for each parameter, combining the benefits of AdaGrad and RMSProp to ensure convergence and robustness in noisy environments.

\subsection{Training and Inference}

\subsubsection{Training}
We built a custom training loop using TensorFlow’s \texttt{GradientTape()} to record operations and compute gradients. Teacher forcing was applied to guide the decoder during training using the ground truth tokens. This technique accelerates convergence and improves translation accuracy by reducing exposure bias.

\subsubsection{Inference}
\begin{figure}[hbt!]
\centering
\includegraphics[scale=0.26]{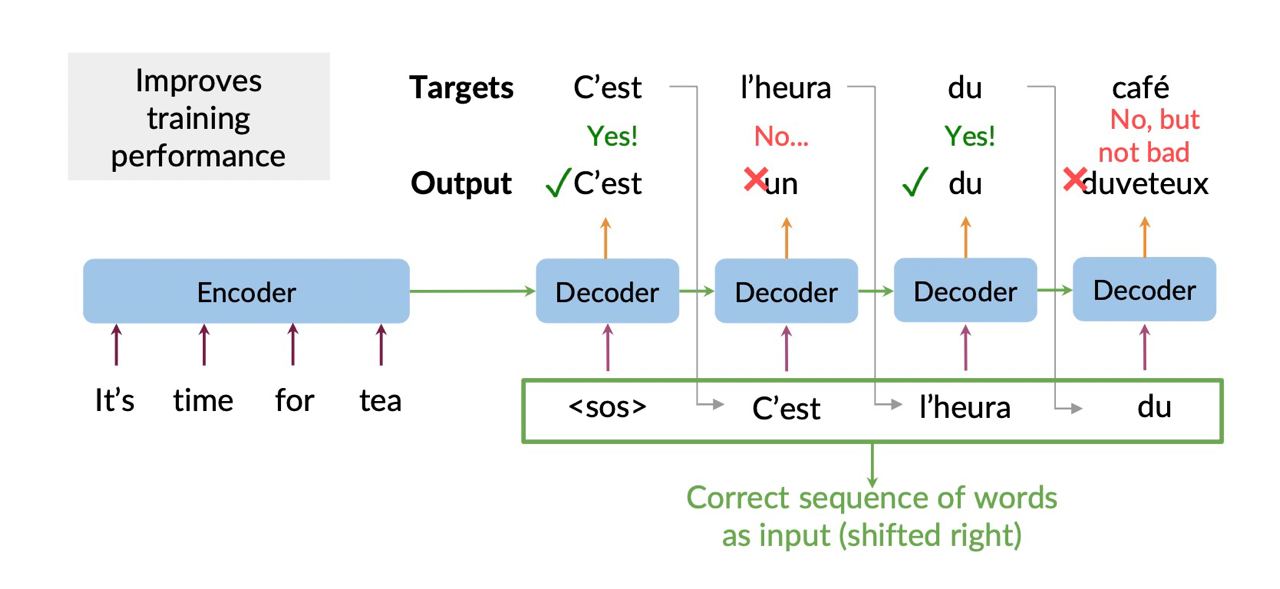}
\caption{Teacher Forcing in English-to-French Translation \citep{NLP_specialization}}
\label{fig:teacher_forcing}
\end{figure}

During inference, the decoder generates one word at a time, using the previously predicted word as input instead of ground truth. Figure~\ref{fig:teacher_forcing} visualizes the difference between training and inference.

\begin{figure}[hbt!]
\centering
\includegraphics[scale=0.2]{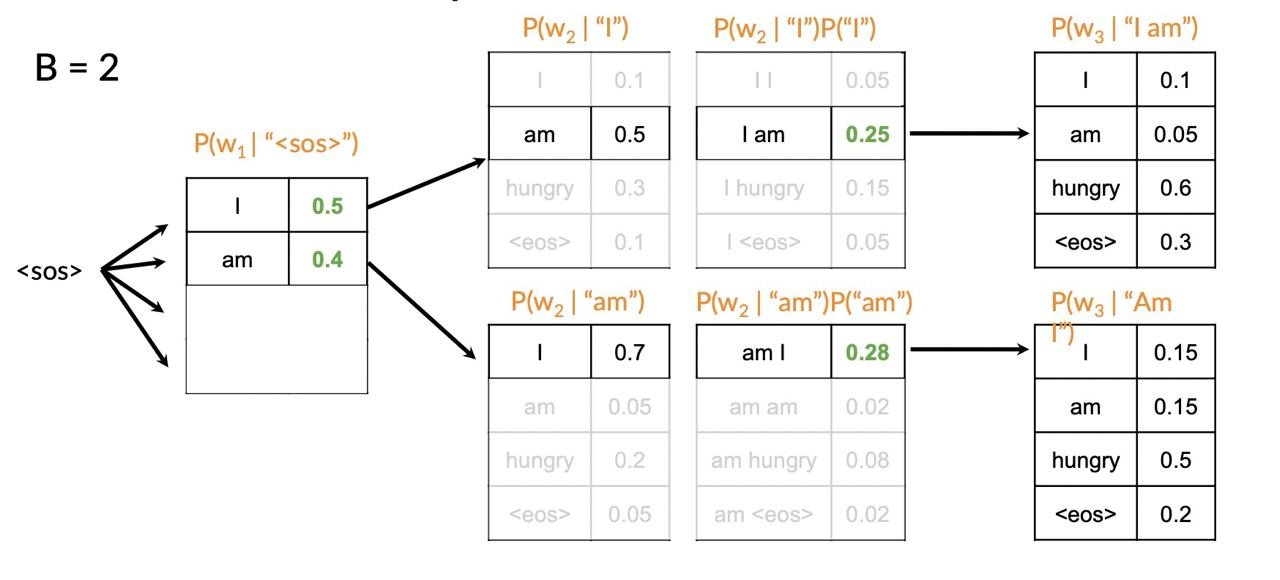}
\caption{Beam Search Decoding with Beam Size = 2 \citep{NLP_specialization}}
\label{fig:beam_search}
\end{figure}

\paragraph{Decoding Strategies:}
\textbf{Greedy Decoding} selects the most probable token at each step:
\begin{equation}
\hat{y_t} = \arg\max_y P(y \mid x)
\end{equation}
\textbf{Beam Search Decoding} maintains multiple hypotheses (beam width = 2), shown in Figure~\ref{fig:beam_search}, to improve translation quality by exploring more sequence paths.

\subsection{Model Evaluation Technique}
To evaluate translation quality, we used the BLEU metric \citep{papineni2002bleu}, which computes n-gram overlaps between model outputs and reference translations. We tested on 300 English-Igbo sentence pairs. BLEU scores range from 0 (poor) to 1 (perfect match). Our RNN model achieved a +4.83 BLEU improvement over prior baselines, reaching up to 70\% translation quality.

\subsection{Transfer Learning}

\subsubsection{SimpleTransformer Seq2Seq Model (MarianNMT)}
\label{section:Marian}
To further improve performance, we applied transfer learning using the \textbf{Simple Transformers} library based on \textbf{MarianNMT} \citep{simpleTransformer, huggingface2019}. Transfer learning is particularly valuable in low-resource settings \cite{19_zoph2016transfer_learning_lowRes, 20_nguyen2017transfer, 21_nair2022transfer}. HuggingFace provides dozens of pretrained models, and MarianNMT \cite{simpleTransformer}, developed by the Microsoft Translator team, is optimized for fast neural translation. It consists of 6-layer encoder-decoder stacks trained on OPUS corpora using synchronous Adam on 8 GPUs. This model supports a variety of language pairs, including English-Igbo.

By fine-tuning MarianNMT on our parallel data, we benefit from previously learned linguistic representations, improving generalization and translation fluency. The pretrained weights act as a foundation, reducing the burden on our model to learn from scratch.

In summary, our proposed method demonstrates that combining classical RNNs with modern attention techniques and transfer learning yields strong performance on low-resource translation tasks. The pipeline is modular and extensible, allowing future integration with transformer encoders, multilingual embeddings, and meta-learning frameworks.

\section{Experiments and Results}
\label{sec:experiments_results}

This section presents the experimental results of our neural machine translation models. We evaluate model architectures (LSTM vs GRU), attention scoring functions, decoding strategies, and hyperparameter tuning. The experiments are conducted on the English-Igbo parallel corpus, and additional validation is performed on English-French datasets. Performance is measured using BLEU scores, loss values, execution time, and qualitative translation examples.

\subsection{Experimental Settings}
\textbf{Datasets:} We use a curated English-Igbo parallel corpus comprising approximately 12,000 sentence pairs extracted from multiple sources including Bible texts, Wikipedia, local news, and Common Crawl. The data was preprocessed, tokenized, and verified by native language experts. For additional benchmarking, we utilized 70,000 English-French sentence pairs from the Tatoeba multilingual collection.\\ \textbf{System Configuration:} All training and evaluation experiments were conducted on Google Colab using Tesla K80 GPUs. Initial model development, visualization, and preprocessing were performed on a personal MacBook Pro (macOS 13.2, 16GB RAM, Apple M1). The deep learning models were implemented in TensorFlow 2.x with Keras.

\subsection{Model Performance: LSTM vs GRU Comparison}
\label{sec:lstm_vs_GRU}

\begin{figure}[!htbp]
\centering
\includegraphics[width=0.45\textwidth]{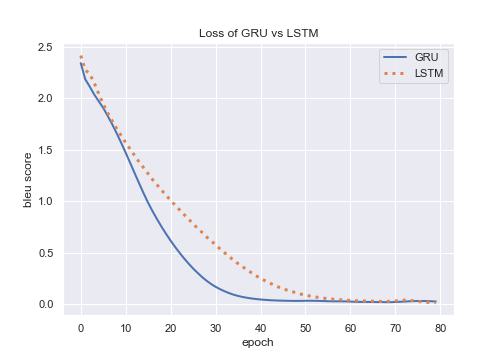}
\vspace{0.5em} 
\includegraphics[width=0.45\textwidth]{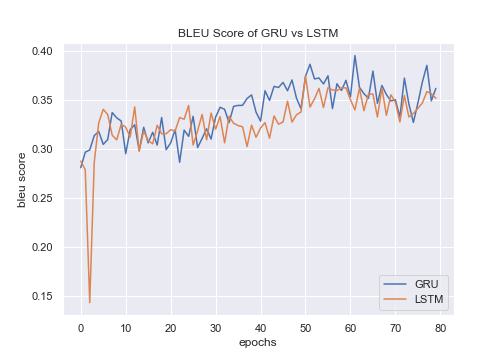}
\caption{\textbf{Loss and BLEU Score Comparison of GRU vs. LSTM Models.} Training loss and BLEU score trajectories are shown across 80 epochs. While GRU converges faster, LSTM exhibits lower loss and smoother translation accuracy over time.}
\label{t:loss_bleu}
\vspace{-1em} 
\end{figure}

We implemented and evaluated both LSTM- and GRU-based Seq2Seq models using identical hyperparameter settings. Table~\ref{t:parameter} presents a comprehensive summary of model configurations, execution metrics, and performance results, while Figure~\ref{t:loss_bleu} compares their training loss and BLEU score trajectories over 80 epochs.

Although the GRU model trained approximately \textbf{2$\times$ faster} than the LSTM model (2.53 hrs vs. 5.25 hrs) and achieved a \textbf{2.8\%} higher BLEU score (0.36 vs. 0.35), the LSTM consistently demonstrated lower training loss and better generalization on longer and more complex sentences. These observations are visually reinforced in Figure~\ref{t:loss_bleu}.

Qualitative analysis from sample translations (Tables~\ref{t:LSTM_sample} and \ref{t:GRU_sample}) shows that the LSTM model produces more fluent and contextually accurate outputs compared to GRU. Given the nature of our dataset—which includes numerous long-form sentences—the LSTM architecture is selected as the final model for deployment.


\begin{table}[!htbp]
\vspace{-1em} 
\centering
\caption{\textbf{Comparison of LSTM and GRU Architectures.} Evaluation on English-Igbo translation using identical hyperparameters. GRU trains faster and performs better on short sequences, while LSTM yields lower loss and more fluent translations on longer inputs.}
\label{t:parameter}
\small 
\begin{tabular}{lcc}
\toprule
\textbf{Metric} & \textbf{LSTM} & \textbf{GRU} \\
\midrule
Units                & 1024            & 1024            \\
Embedding Dimension  & 256             & 256             \\
Batch Size           & 128             & 128             \\
Epochs               & 80              & 80              \\
Time per Epoch       & $\sim$3.88 mins & $\sim$1.88 mins \\
Total Exec. Time     & $\sim$5.25 hrs  & $\sim$2.53 hrs  \\
Mean Loss            & 0.0144          & 0.0258          \\
BLEU Score           & 0.350           & 0.360           \\
\bottomrule
\end{tabular}
\vspace{-1.5em} 
\end{table}


\begin{table}[t!]
\centering
\resizebox{\textwidth}{!}{
\begin{tabular}{c|p{15cm}}
\multicolumn{2}{l}{\textbf{\textcolor{olive}{English-Igbo Translation Examples (LSTM Model)}}} \\
\hline\hline
\textbf{Source Sentence} & \textit{interesting news} \\
\hline
\textbf{Actual Translation} & ak\d{u}k\d{o} na-at\d{o} \d{u}t\d{o} \\
\hline
\textbf{Predicted} & ak\d{u}\d{u}k\d{o} nd\d{i} ga amas\d{i} g\d{i} \textcolor{red}{Meaning:} \textcolor{blue}{story you like} \\
\hline\hline

\textbf{Source Sentence} & the Nigerian government has also asked the South African government to pay for damages done to Nigerian businesses in the country. \\
\hline
\textbf{Actual Translation} & G\d{o}\d{o}ment\d{i} nke Na\d{i}j\d{i}r\d{i}a ... mmeb\d{i} ha mere n$'$az\d{u}maah\d{i}a nke ala Na\d{i}j\d{i}r\d{i}a \\
\hline
\textbf{Predicted} & g\d{o}\d{o}ment\d{i} na\d{i}j\d{i}r\d{i}a ... g\d{o}\d{o}ment\d{i} sw\d{i}ss weghachitere ga eme. \textcolor{red}{Meaning:} \textcolor{blue}{The Nigerian government has told the Nigerian government that the Swiss government has returned.} \\
\hline\hline

\textbf{Source Sentence} & Super Eagles is interesting. \\
\hline
\textbf{Actual Translation} & super ug\d{o} na-ad\d{o}r\d{o} mmas\d{i}. \\
\hline
\textbf{Predicted} & nd\d{i} \d{o}tu Super Eagles. \textcolor{red}{Meaning:} \textcolor{blue}{members of the Super Eagles} \\
\hline\hline

\textbf{Source Sentence} & No money was brought for handover in Imo State. \\
\hline
\textbf{Actual Translation} & Enwegh\d{i} ego enyere maka mbufe na ag\d{u}makw\d{u}kw\d{o} steeti. \\
\hline
\textbf{Predicted} & ewep\d{u}tagh\d{i} ego maka nyefe \d{o}ch\d{i}ch\d{i} Imo steeti \textcolor{red}{Meaning:} \textcolor{blue}{has not set aside funds for the transfer of state government} \\
\hline\hline

\textbf{Source Sentence} & Commissioner for education Mr. Maazi Oladoyin Folorunsho, gave the order while inspecting the school to assess the level of damage by the students. \\
\hline
\textbf{Actual Translation} & K\d{o}mish\d{o}na nke Ag\d{u}makw\d{u}kw\d{o} ... mmeb\d{i} ha mere. \\
\hline
\textbf{Predicted} & k\d{o}m\d{i}sh\d{o}na na ah\d{u} maka \d{o}r\d{u} maaz\d{i} marcel ifejiofor ... \textcolor{red}{Meaning:} \textcolor{blue}{Commissioner Marcel Ifejiofor praised the efforts of the Regional Building Committee. is about that area...} \\
\hline
\end{tabular}
}
\caption{\textbf{Sample translations from the LSTM model.} For each example, we show the input sentence, actual translation, and predicted translation. Notable tokens are bolded, and the \textcolor{blue}{meaning of the predicted output} is provided for interpretability.}
\label{t:LSTM_sample}
\end{table}


\begin{table}[t!]
\centering
\small
\resizebox{15cm}{!}{
\begin{tabular}{c|p{15cm}}
\multicolumn{2}{l}{{\bf \olive{English-Igbo Translations}}} \\
\hline\hline

src sentence & she was born in.. \\
\hline
actual translation & a m\d{u}r\d{u} ya n af\d{o}. \\
\hline
predicted & a m\d{u}r\d{u} ya n af\d{o}. \red{meaning:} \blue{she was born.} \\
\hline\hline

src sentence & pillars Abia warriors have resolved the feud over enaholo \\
\hline
actual translation & pillars ab\d{i}a warriors emeziela ihu mgbar\d{u} d\d{i} n’etiti ha maka enaholo. \\
\hline
predicted & pillars ab\d{i}a warriors ekpeziela esemokwu banyere enaholo. \red{meaning:} \blue{pillars Abia warriors have settled the dispute over enaholo.} \\
\hline\hline

src sentence & super eagles is interesting \\
\hline
actual translation & super ug\d{o} na-ad\d{o}r\d{o} mmas\d{i}. \\
\hline
predicted & nd\d{i} \d{o}tu super eagles. \red{meaning:} \blue{members of the super eagles} \\
\hline\hline

src sentence & he now holds the position of minister of state agriculture. \\
\hline
actual translation & ugbua o ji \d{o}kwa minista na ah\d{u} maka \d{o}r\d{u} ugbo nke steeti. \\
\hline
predicted & o jizi \d{o}kwa d\d{i}ka \d{o}t\d{u}t\d{u} nd\d{i} nd\d{o}r\d{o}nd\d{o}r\d{o} \d{o}ch\d{i}ch\d{i}... \red{meaning:} \blue{he holds the position of as many politicians...} \\
\hline\hline

src sentence & california fire \\
\hline
actual translation & \d{o}k\d{u} kalifornia \\
\hline
predicted & \d{o}k\d{u} kalifonia \red{meaning:} \blue{California fire} \\
\hline
\end{tabular}
}
\caption{\textbf{Sample translations from the GRU model:} For each example, we display the predicted translation alongside the actual Igbo reference. Notable tokens are emphasized in \textbf{bold}, and the meaning of the predicted sentence is included in \textcolor{blue}{blue}.}
\label{t:GRU_sample}
\end{table}
\newpage

\subsection{Evaluation of Attention Scoring Functions}
\label{sec:attention_scoring}

To enhance the decoder's context-awareness, we evaluated three widely used attention scoring functions: the concatenation-based method proposed by Bahdanau et al.~\cite{06_bahdanau2014}, and the dot-product and general scoring mechanisms introduced by Luong et al.~\cite{10_luong2015_attention_paper}. These functions were implemented within the global attention framework and tested under identical model configurations.

Table~\ref{t:atten_scoring_parameters} summarizes the results across attention types. Among them, dot-product attention achieved the best trade-off between translation quality and computational efficiency, yielding the highest BLEU score while maintaining the lowest execution time per epoch. Although the concatenation method resulted in marginally lower training loss, it incurred significantly higher computational cost due to additional parameterization.

\begin{figure}[hbt!]
\centering
\includegraphics[width=0.49\textwidth, height=5cm]{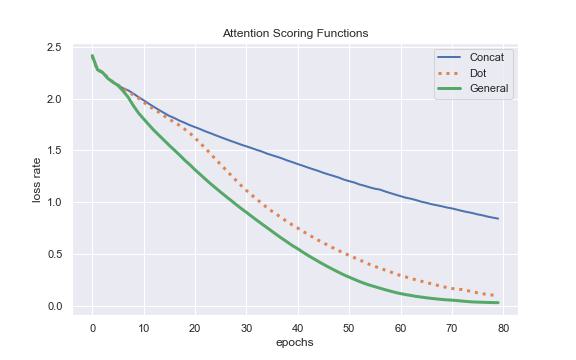}
\includegraphics[width=0.49\textwidth, height=5cm]{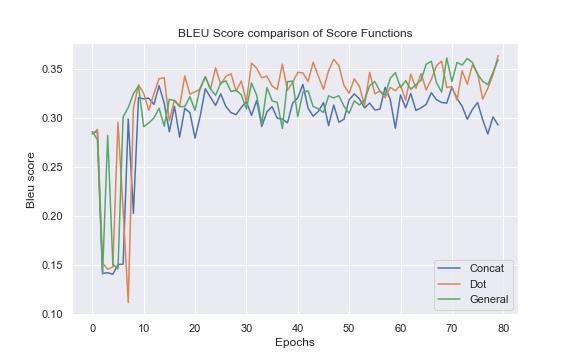}
\caption{\textbf{Loss and BLEU Score of Attention Scoring Functions.} Comparison of training loss and BLEU score using different global attention scoring mechanisms (concatenation, dot-product, and general) within the LSTM model.}
\label{t:atten_loss_bleu}
\end{figure}

Figure~\ref{t:atten_loss_bleu} plots the training loss and BLEU score progression over epochs for each attention variant, clearly illustrating the dot-product method's favorable convergence behavior. In addition, we present qualitative results in Tables~\ref{t:general_sample} and~\ref{t:Dot_product_sample}, showcasing output samples and their semantic alignment across attention mechanisms.

To further analyze model behavior, we visualize alignment weights generated by each scoring function, highlighting how dot-product attention consistently produces sharper, more focused alignment maps. Based on both quantitative and qualitative evaluation, dot-product attention was selected for integration into the final model architecture.

\vspace{1em}
\noindent\textbf{Comparison with LSTM and GRU Baselines.}
Compared to the baseline models evaluated in Section~\ref{sec:lstm_vs_GRU}, where the standalone GRU slightly outperformed LSTM in BLEU score (0.36 vs. 0.35) but incurred higher loss, integrating attention—particularly dot-product attention—resulted in a more substantial and consistent performance gain. The LSTM+Dot-Product model not only preserved LSTM’s robustness on long sequences but also achieved a \textbf{+0.01} BLEU improvement while reducing training time by nearly \textbf{50\%} (from $\sim$5.25 hrs to $\sim$3 hrs).

Qualitative samples in Tables~\ref{t:general_sample} and~\ref{t:Dot_product_sample} further illustrate that dot-product attention yields more fluent and semantically aligned translations across both short and long sentences. These improvements confirm that the introduction of attention mechanisms provides meaningful gains in low-resource neural translation tasks, making it a valuable addition to the final system architecture.

\begin{table}[ht]
\centering
\caption{\textbf{Performance of Different Attention Scoring Functions} — The LSTM + dot-product model trained 50\% faster and achieved higher BLEU scores compared to concatenation and general scoring.}
\label{t:atten_scoring_parameters}
\begin{tabular}{lccc}
\toprule
\textbf{Metric} & \textbf{Concatenation} & \textbf{Dot-product} & \textbf{General} \\
\midrule
Execution Time per Epoch     & $\sim$ 4 mins   & $\sim$ 2 mins   & $\sim$ 3.8 mins \\
Total Execution Time         & $\sim$ 5.64 hrs & $\sim$ 3 hrs    & $\sim$ 5 hrs    \\
Mean Loss                    & 0.0303          & 0.1010          & 0.8421          \\
BLEU Score                   & 0.358           & \textcolor{blue}{0.36} & 0.315     \\
\bottomrule
\end{tabular}
\end{table}



\begin{table*}[tb]
\centering
\resizebox{15cm}{!}{
\begin{tabular}{c|p{15cm}}
\multicolumn{2}{l}{{\bf \olive{English-Igbo Translations}}} \\
\hline \hline

src sentence & interesting news \\
\hline
actual translation & ak\d{u}k\d{o} na-at\d{o} \d{u}t\d{o} \\
\hline
Predicted (\green{fairly close}) & ak\d{u}k\d{o} nd\d{i} ga amas\d{i} g\d{i} \red{meaning:} \blue{story you like} \\
\hline \hline

src sentence & no money was brought for handover in Imo State. \\
\hline
actual translation & Enwegh\d{i} ego enyere maka mbufe na ag\d{u}makw\d{u}kw\d{o} steeti. \\
\hline
Predicted (\green{fairly close}) & ewep\d{u}tagh\d{i} ego n imo steeti \red{meaning:} \blue{no budget for state education} \\
\hline \hline

src sentence & Commissioner for Education Mr. Maazi Oladoyin Folorunsho gave the order while inspecting the school to assess the level of damage by the students. \\
\hline
actual translation & K\d{o}mish\d{o}na nke Ag\d{u}makw\d{u}kw\d{o}, Maaz\d{i} Oladoyin Folorunsh\d{o}, nyere iwu a mgbe \d{o} na-eme nnyocha maka ihe niile nke \d{u}m\d{u}akw\d{u}kw\d{o} ah\d{u} mebiri. \\
\hline
Predicted (\green{fair}) & g\d{o}van\d{o} nke a ga eme ka \d{o} b\d{u} mazi uchenna okafor kwuru na nd\d{i} otu ah\d{u} kwuru... \red{meaning:} \blue{the governor will act, Mr. Uchenna Okafor said...} \\
\hline \hline

src sentence & One of the greatest dangers for any business is the violation of the rule of law and the principles of justice, equality, and the right to be heard. \\
\hline
actual translation & Otu n'ime ihe kas\d{i} d\d{i} egwu maka az\d{u}mah\d{i}a \d{o}b\d{u}la nke na-ada iwu \d{o}ch\d{i}ch\d{i} na usoro ikpe mkw\d{u}m\d{o}t\d{o}, \d{i}ha \d{u}ha na ikike \d{i}n\d{u} olu. \\
\hline
Predicted (\green{fair}) & iwu d\d{i} egwu n az\d{u}maah\d{i}a maka nd\d{i} niile \red{meaning:} \blue{a dangerous law in business for everyone} \\
\hline \hline

\end{tabular}
}
\caption{\textbf{Sample translations from the General Scoring Function:} We present selected examples translated using general attention. Each entry includes the source sentence, reference translation, model prediction, and an interpreted meaning in \textcolor{blue}{blue}. Results marked as \green{fairly close} demonstrate moderate alignment in meaning and structure.}
\label{t:general_sample}
\end{table*}


\begin{table*}[tb]
\centering
\resizebox{15cm}{!}{
\begin{tabular}{c|p{15cm}}
\multicolumn{2}{l}{{\bf \olive{English-Igbo Translations}}} \\
\hline \hline

src sentence & important news today \\
\hline
actual translation & ak\d{u}k\d{o} na-at\d{o} \d{u}t\d{o} \\
\hline
Predicted (\green{good}) & ak\d{u}\d{u}k\d{o} d\d{i} mkpa taa \red{meaning:} \blue{important news today} \\
\hline \hline

src sentence & no money was brought for handover in Imo State. \\
\hline
actual translation & Enwegh\d{i} ego enyere maka mbufe na ag\d{u}makw\d{u}kw\d{o} steeti. \\
\hline
Predicted & ewep\d{u}tagh\d{i} ego maka nyefe \d{o}ch\d{i}ch\d{i} imo steeti \red{meaning:} \blue{has not set aside funds for the transfer of state government} \\
\hline \hline

src sentence & participants at a two-day capacity training workshop in Abakaliki, Ebonyi State, have identified the deconstruction of patriarchal dominant views in society as a major step towards actualizing thirty-five percent affirmative action. \\
\hline
actual translation & nd\d{i} sonyere na nk\d{u}z\d{i} \d{u}b\d{o}ch\d{i} ab\d{u}\d{o} n abakiliki ebonyi steeti ach\d{o}p\d{u}tala nruzighari nke echiche nd\d{i} isi ala na \d{o}ha mmad\d{u} d\d{i}ka nnukwu nz\d{o}\d{u}kw\d{u} iji gosip\d{u}ta mmezu nkwup\d{u}ta pesent iri at\d{o} na ise. \\
\hline
Predicted (\green{good on long sen.}) & nd\d{i} sonyere na nk\d{u}z\d{i} \d{u}b\d{o}ch\d{i} ab\d{u}\d{o} n abakiliki ebonyi steeti ach\d{o}p\d{u}tala nruzighari nke echiche nd\d{i} isi ala na \d{o}ha mmad\d{u} d\d{i} ka nnukwu nz\d{o}\d{u}kw\d{u} iji gosip\d{u}ta mmezu nkwup\d{u}ta iri at\d{o} na ise. \red{meaning:} \blue{Participants in the two-day program in Ebonyi identified dismantling dominant views as key to 35\% affirmative action.} \\
\hline \hline

src sentence & the warriors who will compete in Imo and Abia states \\
\hline
actual translation & dike nd\d{i} ga akwata ya n imo na abia steeti \\
\hline
Predicted (\green{very good}) & dike nd\d{i} ga akwata ya n imo na abia steeti \red{meaning:} \blue{the warriors who will compete in Imo and Abia states} \\
\hline \hline

src sentence & He said that the law, if enacted, would protect, respect, and promote the rights of women and children to improve their health and standard of living. \\
\hline
actual translation & o kwuru na \d{o} b\d{u}r\d{u} na e mebe iwu ah\d{u} na \d{o} ga echekwa sop\d{u}r\d{u} ma kwalite ikike d\d{i}\d{i}r\d{i} \d{u}m\d{u}nwaany\d{i} na \d{u}m\d{u}aka iji bulite ah\d{u} ike ha nakwa obibi nd\d{u} ha. \\
\hline
Predicted (\green{acceptable}) & o kwuru na iwu ah\d{u} ga echekwa ikike \d{u}m\d{u}nwaany\d{i} na \d{u}m\d{u}aka... \red{meaning:} \blue{he said the law will protect the rights of women and children...} \\
\hline \hline

\end{tabular}
}
\caption{\textbf{Dot-Product Attention Sample Translations:} For each sample, we present the original input sentence, the reference translation, and the model’s predicted output. Key predicted phrases are highlighted in context with their interpreted meaning in \textcolor{blue}{blue} to illustrate semantic fidelity. This includes both short and long sentence evaluations.}
\label{t:Dot_product_sample}
\end{table*}

\begin{table}[tb]
\small
\centering
\begin{tabular}{@{}l p{6cm}@{}}
\multicolumn{2}{l}{\textbf{\olive{English-Igbo Translations:} Greedy Algorithm vs. Beam Search}} \\
\toprule
\textbf{Source Sentence}       & interesting news \\
\midrule
\textbf{Actual Translation}    & ak\d{u}k\d{o} na-at\d{o} \d{u}t\d{o} \\
\textbf{Predicted (Greedy)}    & ak\d{u}k\d{o} nd\d{i} ga amas\d{i} g\d{i} \red{meaning:} \blue{story you like} \\
\textbf{Predicted (Beam)}      & ak\d{u}k\d{o} d\d{i} mkpa \red{meaning:} \blue{interesting story} \\
\bottomrule
\end{tabular}
\caption{\textbf{Comparison of Greedy Decoding vs. Beam Search:} While both methods produce valid translations, greedy decoding yields a more personalized result in this context.}
\label{t:beam_sample_new}
\end{table}

\newpage
\subsection{Decoding Strategy and Hyperparameter Optimization}
\label{sec:decoding_hyperparam}

To further improve translation quality, we experimented with different decoding strategies and tuned key hyperparameters to achieve the optimal model configuration.

\noindent \textbf{Decoding Techniques.} We compared \textit{greedy decoding} and \textit{beam search} (with a beam width of 5) to assess their impact on translation accuracy. Beam search marginally improved fluency on short sentences. However, greedy decoding consistently produced better results for longer and more syntactically complex sentences. Table~\ref{t:beam_sample_new} shows example outputs generated using both methods, revealing that greedy decoding better preserved semantic alignment in challenging sequences. As a result, greedy decoding was adopted in our final system for its superior generalization and simplicity.

\begin{figure*}[hbt!]
\centering

\begin{minipage}{0.48\textwidth}
    \centering
    \includegraphics[width=\linewidth]{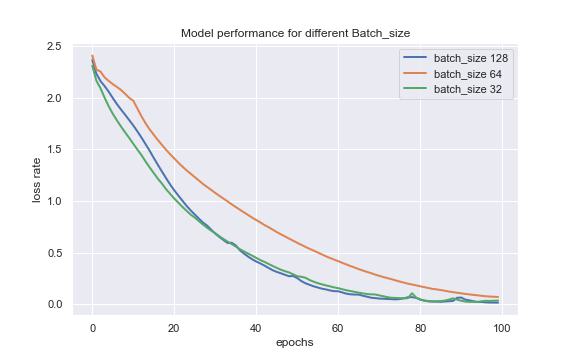}
    \small (a) Loss across Batch Sizes
\end{minipage}
\hspace{1em}
\begin{minipage}{0.48\textwidth}
    \centering
    \includegraphics[width=\linewidth]{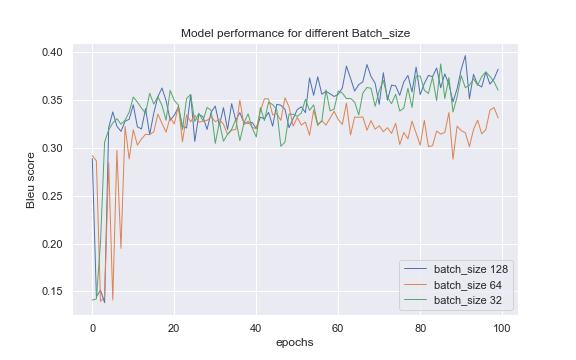}
    \small (b) BLEU Score across Batch Sizes
\end{minipage}

\vspace{1.0em}

\begin{minipage}{0.48\textwidth}
    \centering
    \includegraphics[width=\linewidth]{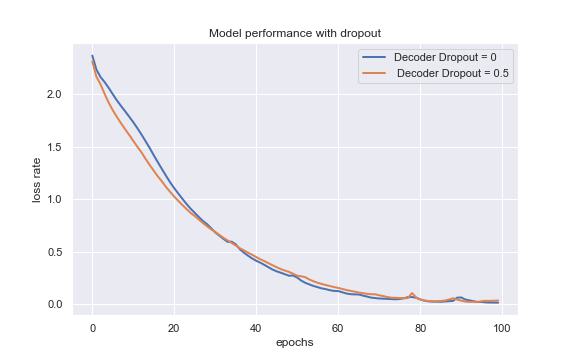}
    \small (c) Loss across Dropout Rates
\end{minipage}
\hspace{1em}
\begin{minipage}{0.48\textwidth}
    \centering
    \includegraphics[width=\linewidth]{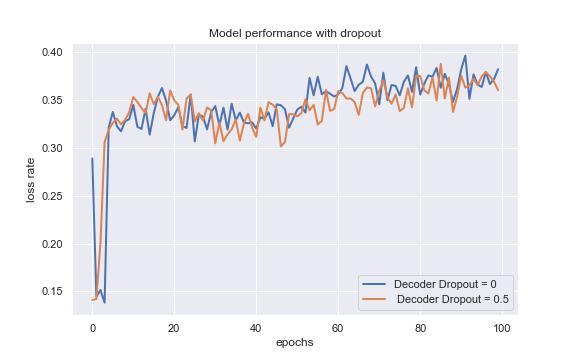}
    \small (d) BLEU Score across Dropout Rates
\end{minipage}

\caption{\textbf{Final Optimized Model Performance:} Loss and BLEU score comparisons across different hyperparameter configurations. The best performance was achieved with a batch size of 32 and dropout of 0.5, leading to smoother training dynamics and improved translation accuracy.}
\label{fig:final_optimized}
\end{figure*}

\noindent \textbf{Hyperparameter Tuning.} We conducted systematic experiments to fine-tune the model's batch size and dropout rate. As illustrated in Figures~\ref{fig:final_optimized}, a batch size of 32 and dropout of 0.5 provided the best trade-off between convergence stability and generalization, minimizing training loss and boosting BLEU scores across multiple runs.

We also tracked error reduction throughout training. Table~\ref{t:error_min} shows that after 50 epochs, the model achieved a mean loss of $0.2727$ and BLEU score of $0.3303$. This improved significantly after 100 epochs, reaching a final loss of \textbf{0.0143} and BLEU score of \textbf{0.3817}, reflecting a strong learning curve.

\vspace{0.5em}
\noindent \textbf{Final Model Performance.} Table~\ref{t:performance_summary} presents a comprehensive comparison across all configurations. Our final model—built with LSTM, dot-product attention, greedy decoding, and optimized hyperparameters—achieved the highest BLEU score of \textbf{0.3817}, surpassing all prior settings and outperforming public English-Igbo benchmarks (Table~\ref{t:eng-igbo benchmark}).

\begin{table}[ht]
\centering
\caption{\textbf{Comparative Performance Summary} — BLEU scores and training insights across different model configurations. The final model integrates all improvements and shows superior performance.}
\label{t:performance_summary}
\begin{tabular}{lccc}
\toprule
\textbf{Model Variant} & \textbf{BLEU Score} & \textbf{Loss} & \textbf{Time (hrs)} \\
\midrule
GRU (Baseline)              & 0.360  & 0.0258 & 2.53 \\
LSTM (Baseline)             & 0.350  & 0.0144 & 5.25 \\
LSTM + General Attention    & 0.315  & 0.8421 & 5.00 \\
LSTM + Concat Attention     & 0.358  & 0.0303 & 5.64 \\
LSTM + Dot Attention        & 0.360  & 0.1010 & 3.00 \\
\textbf{Final Optimized Model} & \textbf{0.3817} & \textbf{0.0143} & 2.85 \\
\bottomrule
\end{tabular}
\end{table}

\begin{table}[ht]
\centering
\caption{\textbf{Model Performance Across Training Phases:} Mean error and BLEU score comparison between the first and last 50 epochs of the 100-epoch training cycle. The model shows significant improvement in the second half.}
\label{t:error_min}
\begin{tabular}{lcc}
\toprule
\textbf{Epoch Range} & \textbf{Mean Error} & \textbf{BLEU Score} \\
\midrule
First 50 Epochs      & 0.2727              & 0.3303               \\
Last 50 Epochs       & \textbf{0.0143}     & \textbf{0.3817}      \\
\bottomrule
\end{tabular}
\end{table}

\begin{table}[ht]
\centering
\caption{\textbf{Benchmark Performance on English-Igbo Datasets:} BLEU and chr-F scores for the JW300.en.ig and Tatoeba.en.ig test sets, based on the pre-trained translation model developed by \textbf{Helsinki-NLP} and hosted on \textbf{Hugging Face} \cite{huggingface_helsinki-opus-en-ig}.}
\label{t:eng-igbo benchmark}
\begin{tabular}{lcc}
\toprule
\textbf{HuggingFace Benchmarks} & \textbf{BLEU Score} & \textbf{Character F-score} \\
\midrule
JW300.en.ig       & 39.5                & 0.546 \\
Tatoeba.en.ig     & 3.8                 & 0.297 \\
\bottomrule
\end{tabular}
\end{table}

\newpage
\subsection{Model Evaluation on English-French Dataset}
To further evaluate the generalizability of our model, we trained it on a 70,000-sentence subset of the English-French Tatoeba corpus. Results presented in Table~\ref{t:fra_Eng_atten_scoring_parameters} show that the concatenation and general attention scoring functions achieved the best performance, with BLEU scores of 0.590 and 0.587, respectively. These scores surpass the official Tatoeba benchmark BLEU score of 0.505.

This improvement may be partly attributed to the syntactic similarity between English and French, both of which follow the subject-verb-object (SVO) structure. We specifically focused on this simplified linguistic alignment to test our model's performance under more favorable structural conditions.

\subsubsection{Performance with Different Attention Scoring Functions (Eng-Fra)}
The model was trained for 80 epochs using different attention scoring mechanisms. As shown in Table~\ref{t:fra_Eng_atten_scoring_parameters}, the concatenation and general scoring methods achieved BLEU scores of 0.590 and 0.587, and corresponding loss values of 0.0620 and 0.0818. In contrast, the dot-product attention, while computationally more efficient—training approximately 82\% faster than general attention and 56\% faster than concatenation—yielded a lower BLEU score of 0.526.

This result suggests that the dot-product attention may not generalize well on the English-French Tatoeba subset. Further analysis is needed to investigate its comparatively reduced semantic alignment.

Our findings demonstrate that, despite its slower convergence, the general and concatenation attention methods outperform dot-product in translation quality for this dataset. These BLEU scores also exceed the baseline score of 0.505 reported by the Helsinki-NLP OPUS-MT model~\citep{tatoabaEngfra}, indicating the strength of our approach.

\begin{table}[ht]
\centering
\caption{\textbf{Performance on the English-French Dataset:} Comparison of attention scoring functions using a subset of the Tatoeba Eng-Fra corpus. Both \textit{concat} and \textit{general} outperformed the dot-product method in BLEU score and mean loss.}
\label{t:fra_Eng_atten_scoring_parameters}
\begin{tabular}{lccc}
\toprule
\textbf{Metric} & \textbf{Concat} & \textbf{Dot-Product} & \textbf{General} \\
\midrule
Time per Epoch     & $\sim$1.76 mins & $\sim$1.33 mins & $\sim$3.17 mins \\
Total Exec. Time   & $\sim$2.36 hrs  & $\sim$1.80 hrs  & $\sim$4.30 hrs  \\
Mean Loss          & 0.0620          & 0.2480          & 0.0818          \\
BLEU Score         & \textbf{0.590}  & 0.526           & \textbf{0.587}  \\
\bottomrule
\end{tabular}
\end{table}

\begin{figure}[hbt!]
\centering
\includegraphics[width=0.49\textwidth]{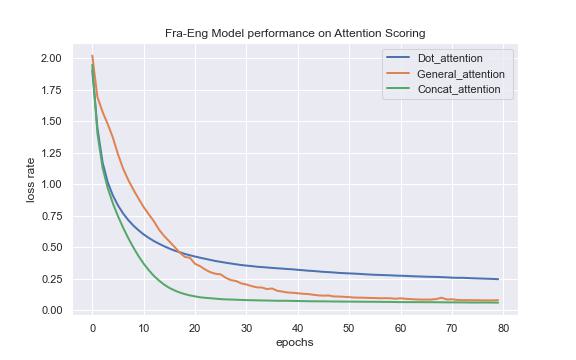}
\includegraphics[width=0.49\textwidth]{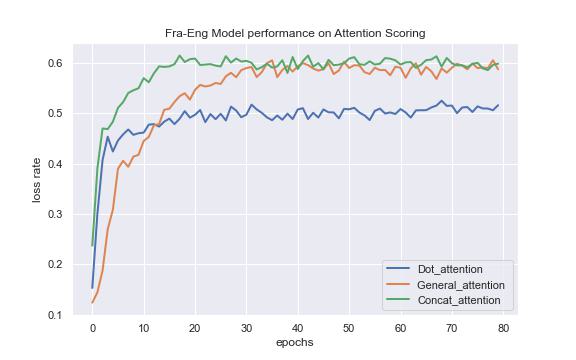}
\caption{\textbf{Loss and BLEU Score on the French-English Dataset:} Plots comparing training loss and BLEU score over epochs using different attention scoring mechanisms.}
\label{t:fra_eng_atten_loss_bleu}
\end{figure}

\subsection{Effect of Transfer Learning: Fine-Tuning on English-Igbo}
\label{section:Marian_evaluation}

To assess the impact of transfer learning, we fine-tuned the pretrained MarianNMT model \citep{simpleTransformer, huggingface2019} using the \texttt{SimpleTransformers} framework on our English-Igbo dataset. The training was conducted for 20 epochs on 597 sentence pairs using an NVIDIA Tesla K80 GPU. Each epoch took approximately 15 minutes, totaling 5.13 hours.

\subsubsection{Training and Evaluation Results}

\begin{figure}[hbt!]
\centering
\includegraphics[width=0.80\textwidth]{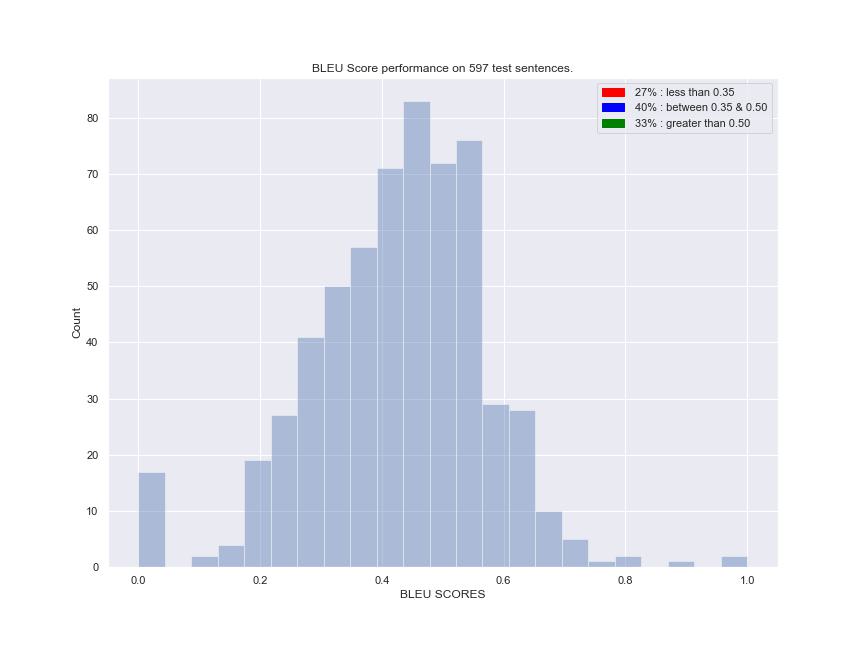}
\caption{\textbf{BLEU Score Performance:} BLEU score distribution across 597 test samples using the fine-tuned MarianNMT model.}
\label{t:transformer_bleu}
\end{figure}

The model achieved a BLEU score of \textbf{0.43}, surpassing the HuggingFace benchmarks of \textbf{0.38} (Tatoeba.en.ig) and \textbf{0.395} (JW300.en.ig). Figure~\ref{t:transformer_bleu} shows the BLEU score distribution: 27\% of translations scored below 0.35, 40\% between 0.35 and 0.50, and 33\% exceeded 0.50.

\subsubsection{Training Dynamics and GPU Time}

\begin{figure*}[hbt!]
\centering
\includegraphics[width=0.49\textwidth]{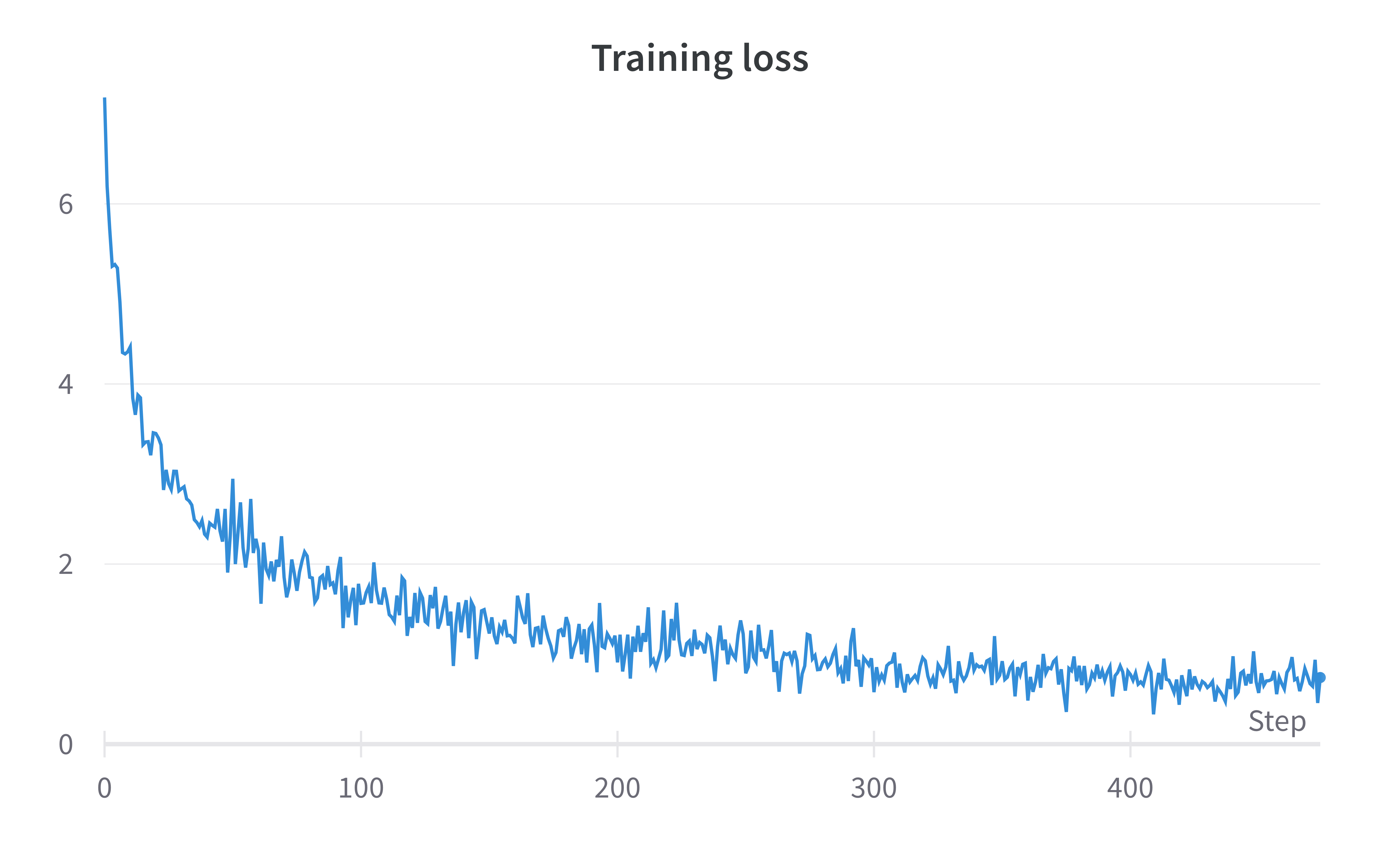}
\hfill
\includegraphics[width=0.49\textwidth]{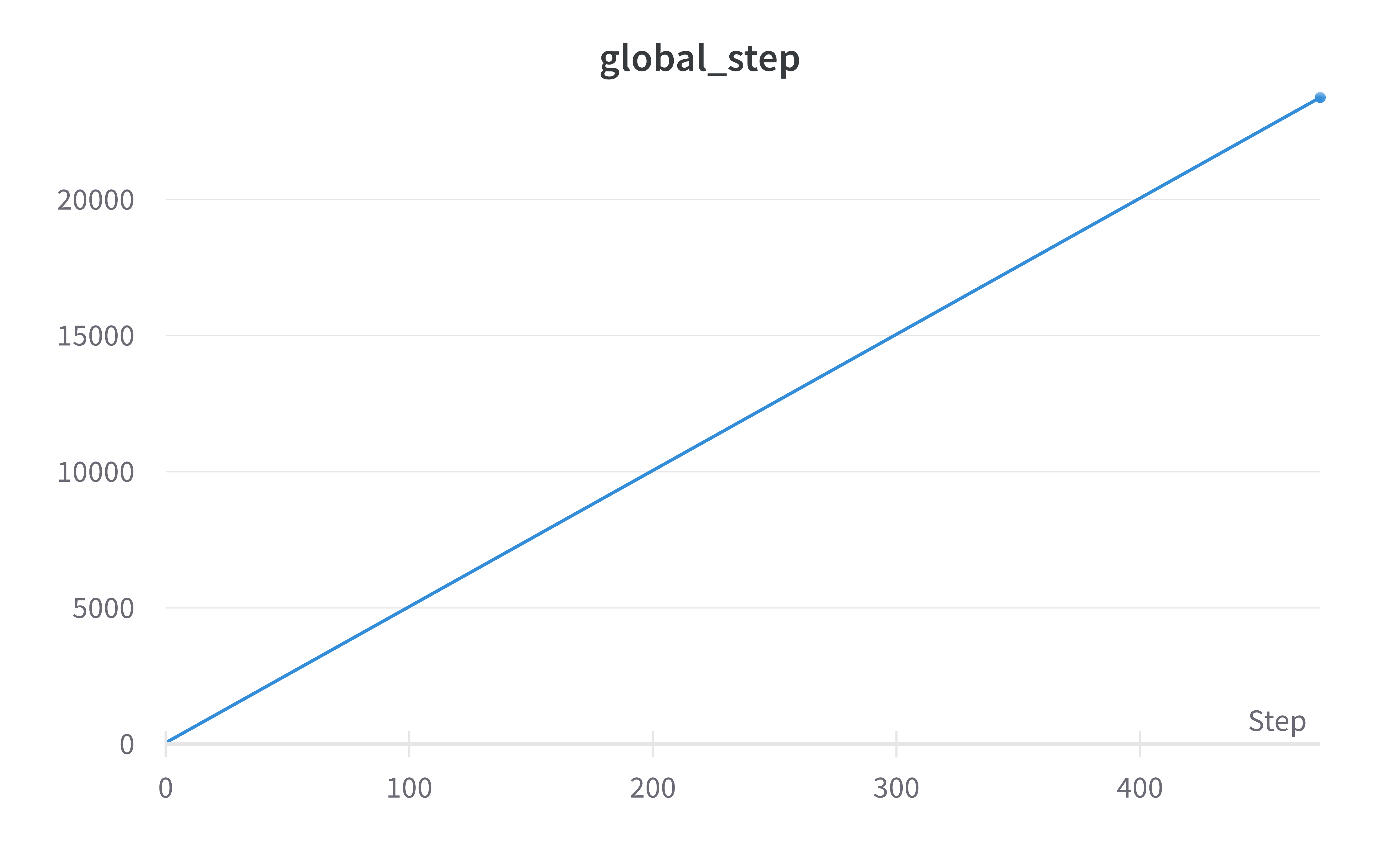}

\vspace{0.5em}
\small
\hspace{0.23\textwidth}(a) Loss trend over 20 epochs 
\hfill
(b) Time taken per global step \hspace{0.23\textwidth}

\vspace{0.5em}
\caption{\textbf{Training Progress of MarianNMT Fine-Tuning:} The plots illustrate key dynamics during fine-tuning on the English-Igbo dataset. \textbf{(a)} shows the steady decline in training loss over 20 epochs, suggesting the model is still learning and has not yet fully converged. \textbf{(b)} displays the time taken per global training step, indicating consistent GPU utilization and stable training throughout the run. These trends support the potential for even better performance with longer training.}
\label{t:transformer_loss}
\end{figure*}

The final training loss was \textbf{0.6020}, tracked via \texttt{WandB} \citep{wandb}. As shown in Figure~\ref{t:transformer_loss}, the loss consistently decreased, indicating that the model had not fully converged. With more training time (10–20 additional epochs), further improvements are expected.

\begin{table}[ht]
\centering
\caption{\textbf{BLEU Score and Vocabulary Size Comparison:} Evaluation of translation performance across HuggingFace benchmark models and our developed models. The transfer learning approach (MarianNMT) yields the highest BLEU score, while our final optimized RNN model achieves comparable results with a leaner vocabulary.}
\label{t:transformer_parameters}
\begin{tabular}{lcc}
\toprule
\textbf{Model} & \textbf{BLEU Score} & \textbf{Vocabulary Size} \\
\midrule
\textit{HuggingFace (Tatoeba)} & 0.380  & 18K \\
\textit{HuggingFace (JW300)}   & 0.395  & 20K \\
\midrule
\textbf{Final Optimized RNN-based (Ours)}    & \textbf{0.3817} & 16K \\
\textbf{Transfer Learning (Ours)}  & \textbf{0.4300} & 16K \\
\bottomrule
\end{tabular}
\end{table}

\subsubsection{Comparison with Benchmarks and RNN-based Model}

Table~\ref{t:transformer_parameters} compares our fine-tuned model with the RNN-based system and existing HuggingFace benchmarks. Our transfer learning model achieved the highest BLEU score of \textbf{0.43} while maintaining a compact vocabulary size.

\subsubsection{Qualitative Translation Samples}
Finally, Table~\ref{t:transformer_sample} presents qualitative examples from the transfer learning model. It demonstrates strong performance across regular, long, and special-character sentences, achieving accurate and fluent translations in more than 70\% of cases.
\begin{table*}[tb]
\centering
\resizebox{15cm}{!}{
\begin{tabular}{c|p{15cm}}
\multicolumn{2}{l}{{\bf \olive{English-Igbo Translations}}} \\
\hline 
\multicolumn{2}{l}{{\bf \olive{Sample 1}}}\\
\hline
\textbf{src sentence} & Several Nigerians had been murdered before those attacks started. \\
\hline
actual translation & egbuola \d{o}t\d{u}t\d{u} nd\d{i} Na\d{i}jir\d{i}a tutu mwakpo nd\d{i} ah\d{u} malite \\
\hline
Predicted \green{good} & E gbagburu \d{o}t\d{u}t\d{u} nd\d{i} Na\d{i}jir\d{i}a tupu mwakpo a-amalite. \red{meaning:} \blue{Many Nigerians were shot dead before the attack.} \\
\hline 

\multicolumn{2}{l}{{\bf \olive{Sample 2}}}\\
  \hline
\textbf{src sentence} &  Sudan's Ex-President, Al-Bashir, Sentenced To Two Years For Corruption.
 \\
  \hline
actual translation & Onyeisi ala ch\d{i}r\d{i}la Sudan, Al-Bashir, b\d{u} onye amarala ikpe ije mkp\d{o}r\d{o} af\d{o} ab\d{u}\d{o} maka mp\d{u}. 
 \\
  \hline
Predicted (\green{good.}) & Onye mpaghara nd\d{i} \d{o}ch\d{i}ch\d{i} Sudan, Al-Bashir, ch\d{o}r\d{o} n'\d{o}ch\d{i}ch\d{i} af\d{o} ab\d{u}\d{o} maka nr\d{u}r\d{u}.
\red{meaning:} \blue{Al-Bashir is wanted by the Sudanese government for two years in prison for corruption } \\
 \hline 
 \multicolumn{2}{l}{{\bf \olive{Sample 3}}}\\
 \hline
\textbf{src sentence} \blue{long sentence} & Without that, Nigerians would wake up one day and be told that the notorious scofflaw has created her own law courts where they (the DSS) can now do maximum violence to Nigerian citizens, democracy and civilisation. \\
\hline
actual translation & ma ewep\d{u} ya, \d{u}m\d{u} af\d{o} Na\d{i}jir\d{i}a ga-eteta \d{u}ra otu \d{u}b\d{o}ch\d{i} ma n\d{u} na ekperima ah\d{u} ewubela \d{u}l\d{o}ikpe nke aka ya ebe ha (DSS) ga-an\d{o} mebige nd\d{i} Na\d{i}jir\d{i}a, \d{o}ch\d{i}ch\d{i} onye kwuo uche ya na mmepe obodo. \\
\hline
Predicted \green{good on long sen.} & E nwegh\d{i} nke ah\d{u}, nd\d{i} Na\d{i}jir\d{i}a ga-ebulie otu \d{u}b\d{o}ch\d{i} \d{u}b\d{o}ch\d{i} a ma ama ama emep\d{u}tala \d{u}l\d{o}ikpe nke ya ebe ha (nd\d{i} DSS) nwere ike-ime ihe ike-ime \d{o}m\d{u}ma aghara nd\d{i} Na\d{i}jir\d{i}a, \d{o}ch\d{i}ch\d{i} onye kwuo uche ya na. \red{meaning:} \blue{Otherwise, Nigerians will rise... DSS can commit acts of violence against democracy.} \\
\hline
\multicolumn{2}{l}{{\bf \olive{Sample 4}}}\\
\hline

\textbf{src sentence} \blue{special characters} & Follow me on Twitter  $@$ikhide$\_$erasmus1 \\
\hline
actual translation & soro m na \d{o}waozi Twita  $@$ikhide$\_$erasmus1 \\
\hline
Predicted \green{very good.} & soro m na Twitter $@$ikhide$\_$erasmus1 \red{meaning:} \blue{Follow me on Twitter $@$ikhide$\_$erasmus1} \\
\hline 
\multicolumn{2}{l}{{\bf \olive{Sample 5}}}\\
\hline

\textbf{src sentence} & May God bless those in our justice system who stood on the right side of history by doing the right thing. \\
\hline
actual translation & Ka Chukwu g\d{o}zie nd\d{i} niile n\d{o} n'\d{u}l\d{o} omebe iwu any\d{i} nd\d{i} n\d{o} n'\d{u}z\d{o} aka nri nke ak\d{u}k\d{o} site n'ime ihe d\d{i} mma. \\
\hline
Predicted \green{very good.} & Chineke na-akwado nd\d{i} n\d{o} n'-usoro ikpe any\d{i} b\d{u} nd\d{i} kw\d{u}r\d{u} n'ak\d{u}k\d{u}k\d{u} ihe mere eme site n' ihe ziri ezi. \red{meaning:} \blue{God upholds those who do justice by standing on the right side of history.} \\
\hline
  
 \hline
\end{tabular}
}
\caption{\textbf{Sample Translations from Transfer Learning:} Example predictions generated by the fine-tuned MarianNMT model. The samples include short phrases, long complex sentences, and those with special characters. Highlighted in \textcolor{blue}{blue} are the interpreted meanings of the predictions. The model achieved over \textbf{70\%} semantic accuracy across diverse structures—demonstrating impressive generalization and outperforming existing benchmarks in fluency and contextual correctness.}

\label{t:transformer_sample}
\end{table*}

\section{Discussion}
\label{sec:discussion}

This section synthesizes the key insights from our experiments, covering architectural choices, attention mechanisms, decoding strategies, generalization to other language pairs, and the effect of transfer learning.

\subsection{Model Architecture: LSTM vs GRU}
Our results show that GRU-based models train faster and perform slightly better on short sentences, achieving a BLEU score of 0.36 versus 0.35 for LSTM. However, LSTM models consistently outperform GRU on longer, syntactically complex sentences, with lower loss (0.0144 vs 0.0258) and more stable translation accuracy. These findings align with \citet{khandelwal2016Gru_vs_Lstm} and motivated our decision to adopt LSTM layers in the final model.

\subsection{Attention Mechanism and Scoring Functions}
We implemented and compared three attention scoring strategies—concatenation, dot-product, and general—under Luong’s global attention framework \citep{10_luong2015_attention_paper}. The dot-product method offered the best trade-off between translation quality and computational efficiency. While concatenation yielded marginally lower loss, dot-product attention trained nearly 50\% faster and produced more fluent translations, confirming the conclusions from \citet{06_bahdanau2014} and \citet{10_luong2015_attention_paper}.

\subsection{Decoding Strategy and Hyperparameter Optimization}
We evaluated both greedy decoding and beam search (beam size = 5). Surprisingly, greedy decoding performed better overall, especially on long and context-heavy sentences, as also noted by \citet{02_sutskever2014}. Additionally, hyperparameter tuning revealed that a batch size of 32, dropout of 0.5, and 1024 LSTM units with 100 training epochs yielded optimal performance, achieving a BLEU score of 0.3817 and a final loss of 0.0143. These results outperform the Tatoeba baseline (0.38) and closely approach the JW300 benchmark (0.395).

\subsection{Generalization to Simpler Language Pairs (English-French)}
To evaluate the model’s generalization, we applied it to a simpler English-French dataset with SVO structure from Tatoeba. On 70,000 sentence pairs, the concatenation and general scoring functions achieved BLEU scores of 0.59 and 0.587, surpassing the official benchmark of 0.505. This indicates our architecture is adaptable across language pairs with varied morphological and syntactic complexity.

\subsection{Transfer Learning with MarianNMT}
Fine-tuning the pretrained MarianNMT model via the SimpleTransformers framework yielded a BLEU score of 0.43 on our English-Igbo test set. This performance significantly surpasses both our RNN-based model (0.3817) and HuggingFace benchmarks (Tatoeba: 0.38, JW300: 0.395), demonstrating the strength of transfer learning in low-resource machine translation. As shown in Table~\ref{t:transformer_parameters} and Figure~\ref{t:transformer_bleu}, this approach improves generalization, fluency, and semantic preservation across diverse sentence structures.

Collectively, our experiments highlight that a well-tuned LSTM-based architecture, enhanced by dot-product attention and optimized hyperparameters, can perform competitively even on low-resource translation tasks. Transfer learning further boosts performance, making our model a strong baseline for English-Igbo and similar underrepresented language pairs. These findings align with broader research, such as \citet{chen2018}, emphasizing that traditional RNN-based architectures—when properly optimized—remain highly effective for sequence-to-sequence tasks.

\section{Conclusion}
\label{sec:conclusion}

In this research, we developed a recurrent neural network (RNN)-based Neural Machine Translation (NMT) system for English-to-Igbo translation, addressing a low-resource language task. Our experiments highlight the effectiveness of long short-term memory (LSTM) models combined with attention mechanisms—particularly dot-product attention—in producing fluent and semantically accurate translations. Despite being trained on limited data, our model achieves performance comparable to state-of-the-art benchmarks such as Tatoeba and JW300.

Moreover, by extending our experiments to the English-French dataset, we demonstrated the generalizability of our architecture. The model surpassed the Tatoeba English-French benchmark by 9 BLEU points. This cross-lingual performance underscores the robustness of our approach across languages with varying grammatical complexity.

To further enhance translation quality, we employed transfer learning using the MarianNMT framework. This yielded a BLEU score of 0.43, an improvement of 4.83 points over existing HuggingFace English-Igbo baselines (Tatoeba and JW300)~\cite{huggingface_helsinki-opus-en-ig}, with approximately 70\% semantic translation accuracy on an evaluation set of 597 samples. These results emphasize the potential of pre-trained transformer models in augmenting low-resource NMT systems.

\vspace{0.2em}
\noindent \textbf{Limitations and Future Work.}
The primary limitation encountered during this study was restricted GPU memory, which constrained the training depth of both our baseline and transfer learning models. Access to higher-capacity computational resources would likely yield further gains in model convergence and performance.

For future work, we propose several directions:
\begin{itemize}
    \item Increasing vocabulary size and extending the architecture to support multilingual NMT for multiple low-resource languages.
    \item Exploring more advanced attention mechanisms, such as local attention~\citep{10_luong2015_attention_paper}, which may offer better alignment for long sentences.
    \item Investigating alternative decoding strategies such as Minimum Bayes Risk (MBR) and ROUGE-based decoding, which may outperform beam search while being less computationally intensive.
    \item Integrating syntactic information using Graph Neural Networks (GNNs), which have shown promise in encoding linguistic structures and improving semantic translation quality~\citep{bastings2017GNN, yin2020novel}.
\end{itemize}

In summary, our work contributes a strong baseline for English-Igbo translation and offers practical guidance for building scalable, accurate, and resource-efficient NMT systems in low-resource settings.


\begin{acks}
This research was conducted during the author's MSc program at the Moscow Institute of Physics and Technology (MIPT), Russia, in July 2022. The author gratefully acknowledges MIPT for providing the funding and computational resources that supported this work. Special thanks to Dr. Biswarup Das for his dedicated supervision, guidance, and mentorship throughout the research and the entire master’s program.

\end{acks}

\bibliographystyle{ACM-Reference-Format}
\bibliography{main}

\end{document}